\documentclass[conference]{IEEEtran}
\IEEEoverridecommandlockouts

\usepackage{fontspec}

\newfontfamily\cyrillicfont{DejaVuSerif}[
  Extension      = .ttf,
  UprightFont    = *,
  BoldFont       = *-Bold,
  ItalicFont     = *-Italic,
  BoldItalicFont = *-BoldItalic,
]
\newcommand{\cyr}[1]{{\cyrillicfont #1}}

\usepackage{xeCJK}

\setCJKsansfont{FandolHei-Regular.otf}[
  BoldFont = FandolHei-Bold.otf,
]
\setCJKmonofont{FandolFang-Regular.otf}

\usepackage{cite}
\usepackage{amsmath,amssymb,amsfonts}
\usepackage{algorithmic}
\usepackage{graphicx}
\usepackage{textcomp}
\usepackage{xcolor}
\usepackage{microtype}
\usepackage{booktabs}
\usepackage{multirow}
\usepackage{hyperref}
\usepackage{tabularx}
\usepackage{array}

\def\BibTeX{{\rm B\kern-.05em{\sc i\kern-.025em b}\kern-.08em
    T\kern-.1667em\lower.7ex\hbox{E}\kern-.125emX}}

\newcolumntype{L}[1]{>{\raggedright\arraybackslash}p{#1}}
\newcolumntype{C}[1]{>{\centering\arraybackslash}p{#1}}
\newcolumntype{R}[1]{>{\raggedleft\arraybackslash}p{#1}}

\begin{document}

\title{Efficient Multilingual Name Type Classification Using Convolutional Networks}

\author{\IEEEauthorblockN{Davor Lauc}
\IEEEauthorblockA{\textit{University of Zagreb / \href{https://mondonomo.ai}{Mondonomo AI}} \\
Zagreb, Croatia / Wilmington, DE, US \\
dlauc@ffzg.unizg.hr; \href{mailto:davor@mondonomo.ai}{davor@mondonomo.ai}}}

% IEEE Copyright Notice (bottom-left of page 1)
\IEEEpubid{\makebox[\columnwidth]{979-8-3315-0217-1/25/\$31.00~\copyright~2025 IEEE \hfill}%
\hspace{\columnsep}\makebox[\columnwidth]{ }}

\maketitle
\IEEEpubidadjcol

\begin{abstract}
We present a convolutional neural network approach for classifying proper names by language and entity type. Our model, Onomas-CNN X, combines parallel convolution branches with depthwise-separable operations and hierarchical classification to process names efficiently on CPU hardware. We evaluate the architecture on a large multilingual dataset covering 104 languages and four entity types (person, organization, location, other). Onomas-CNN X achieves 92.1\% accuracy while processing 2,813 names per second on a single CPU core—46 times faster than fine-tuned XLM-RoBERTa with comparable accuracy. The model reduces energy consumption by a factor of 46 compared to transformer baselines. Our experiments demonstrate that specialized CNN architectures remain competitive with large pre-trained models for focused NLP tasks when sufficient training data exists.
\end{abstract}


\section{Introduction}
Proper names appear in virtually every text document but remain understudied in computational linguistics \cite{lauc2024}. This gap creates practical problems for multilingual NLP systems that must process names of people, places, organizations, and other entities across different languages and scripts.

The challenges are significant. Names are sparse in standard dictionaries and corpora. The same entity may have dozens of valid spellings across languages—consider how ``Alexander'' can be transliterated as ``\cyr{Александер}'' or adapted as the native form ``\cyr{Александр}'' in Russian, ``アレクサンダー'' in Japanese, and ``亚历山大'' in Chinese. Some Thai names have dozens of observed transliterations \cite{lauc2024ayutthaya}. Additionally, new names emerge constantly as companies form, people gain prominence and migrate, and language evolves.

Current resources provide limited coverage. JRC-Names contains approximately 340,000 entities with their name variants \cite{jrc2013}, while ParaNames offers 16.8 million entities with 140 million name variants across 400+ languages \cite{saleva2024}. However, even these large collections miss many names found in real-world text, particularly for personal and organizational entities outside encyclopedic sources.

Recent benchmarks reveal that state-of-the-art multilingual models struggle with name processing, especially in low-resource languages \cite{mayhew2024}. The Universal NER dataset shows performance drops of 15-30\% when moving from high-resource to low-resource languages. From a linguistic perspective, names carry information about language contact, cultural exchange, and historical relationships \cite{mignot2022}. They serve as markers of identity and reveal patterns of migration and cultural influence.

\subsection{The Ambiguity Problem}
Name ambiguity creates fundamental challenges for automated systems. The same text string often represents different entity types depending on context. Consider ``Ford''—it could mean Henry Ford (person), Ford Motor Company (organization), or a Ford vehicle (product). Research indicates that approximately 19\% of European and U.S. firms are eponymous, named after their founders \cite{belenzon2017}. This systematic ambiguity affects millions of entity mentions in text.

Geographic names pose similar problems. Many surnames derive from place names \cite{shaw2024}. ``Washington'' might refer to George Washington, Washington D.C., or the state of Washington. Medieval records show toponymic surnames indicated residence, ancestry, or migration patterns. Modern NLP systems must disambiguate these overlapping meanings. Nested entities add another layer of complexity \cite{ling2015}. ``New York Times'' contains both a location (New York) and an organization (the newspaper). Entity linking systems must handle these competing interpretations, often requiring contextual information to resolve ambiguities correctly.

\subsection{Cross-Linguistic Challenges}
Multilingual name processing faces additional obstacles beyond simple translation. Names undergo complex transformations when crossing language boundaries. Transliteration between scripts lacks standardization \cite{merhav2018}. The same name may have multiple valid representations in another script, each emphasizing different phonetic or orthographic aspects. Chinese media renders ``Trump'' as both ``川普'' (Chuānpǔ) and ``特朗普'' (Tèlǎngpǔ), reflecting regional preferences rather than errors.

Cultural naming practices vary widely \cite{heffernan2010, kim2017}. East Asian professionals often use Western names in international contexts. Arabic names follow patronymic patterns unfamiliar to Western-trained systems. Indigenous naming traditions may include meaningful phrases that standard tokenizers incorrectly split. Brand names face localization decisions \cite{delacova2021}. Some companies maintain uniform English names globally (Google, Microsoft), while others adapt to local markets (Volkswagen becomes ``大众'' in Chinese, meaning ``the masses'').

\subsection{Limitations of Current Approaches}
Transformer-based models like XLM-RoBERTa and mBERT have become the default choice for multilingual NLP tasks. However, these models face practical deployment challenges. Their large size (hundreds of millions of parameters) requires GPU acceleration for acceptable throughput. Energy consumption is substantial, with a single inference pass consuming significantly more power than lighter architectures. Memory requirements often exceed commodity hardware specifications. For production systems processing millions of names daily, these constraints translate to high operational costs and environmental~impact.

Fine-tuning large pre-trained models on name classification tasks yields strong accuracy but does not address efficiency concerns. The self-attention mechanism processes all positions quadratically, creating bottlenecks even for short inputs. Quantization and distillation techniques reduce size but still lag behind purpose-built architectures in throughput.

\subsection{Our Approach}
This paper develops an efficient CNN-based approach to multilingual name classification. We make three main contributions:

First, we design Onomas-CNN X, a convolutional architecture optimized for CPU deployment. The model uses depthwise-separable convolutions and hierarchical classification to achieve high throughput without GPUs.

Second, we conduct extensive experiments comparing CNN variants against transformer baselines. We evaluate accuracy, speed, calibration, and energy consumption across diverse test conditions.

Third, we provide practical deployment guidelines for production systems, including optimal batch sizes and quantization strategies for different hardware configurations.

Our results challenge the assumption that larger models always perform better. For focused tasks with sufficient training data, specialized architectures can match transformer performance while using far less computational resources.

\section{The Proposed Method}
We design Onomas-CNN X specifically for efficient name classification on CPU hardware. The architecture balances accuracy with inference speed through several key innovations.

\subsection{Architecture Components}
The model consists of four main components working in sequence.

\textbf{Embeddings:} We use 384-dimensional learned embeddings for token-level input. Positional encodings capture character order, which matters for morphological patterns in many languages. We apply 5\% dropout during training to prevent overfitting.

\textbf{Parallel Convolutions:} Five branches process the input simultaneously with kernel sizes 1 through 5. Each captures different pattern scales: size 1 for individual characters and diacritics, size 2 for bigrams and digraphs, size 3 for common morphemes, and sizes 4-5 for longer substrings and compound elements. Filter counts follow the formula $128 \times 1.5^{(i \times 0.5)}$ for branch $i$, allocating more capacity to branches capturing larger patterns.

\textbf{Pooling Strategy:} We combine three pooling methods with learned weights:
\begin{equation}
\text{output} = w_1 \cdot \text{max\_pool} + w_2 \cdot \text{avg\_pool} + w_3 \cdot \text{att\_pool}
\end{equation}
Max pooling captures distinctive features, average pooling provides global context, and attention pooling learns position-specific importance. The model learns optimal weights through backpropagation.

\textbf{Hierarchical Classifier:} Instead of 476 direct classes, we use two stages: first classifying into 24 language clusters, then classifying within each cluster. This reduces complexity from O(476) to O(24 + 20) and exploits linguistic relationships between languages.

\subsection{Efficiency Optimizations}
Several design choices improve CPU performance:

\textbf{Depthwise Separable Convolutions:} Following Mobile\-Nets~\cite{howard2017}, we decompose standard convolutions into depthwise and pointwise operations. With 8~groups, this reduces parameters substantially compared to standard convolutions. The exact reduction depends on filter configurations but typically ranges from 8$\times$ to~9$\times$.

\textbf{Memory Layout:} The 338 MB model fits in L3 cache on modern CPUs, avoiding main memory access. Parallel branches reuse input embeddings efficiently. The hierarchical classifier only loads weights for the predicted language cluster.

\textbf{Quantization Support:} The architecture works well with INT8 quantization. ReLU activations and convolution operations maintain accuracy under reduced precision. We observe only 0.3\% accuracy loss with 1.8x speedup.

\subsection{Training Procedure}
We address class imbalance through careful training. The loss function uses focal loss with class weights to handle the imbalanced distribution:
\begin{equation}
FL(p_t) = -\alpha_t(1 - p_t)^{\gamma} \log(p_t)
\end{equation}
We use $\alpha = 0.25$ and $\gamma = 2.0$ to focus on hard examples. Progressive training proceeds in three stages: train the language cluster classifier alone, freeze it and train entity classifiers, then fine-tune the complete model jointly. Data augmentation applies capitalization changes (30\%), title and abbreviation modifications (20\%), and character noise (10\%) during training.

\subsection{Alternative Architectures}
We also test three CNN variants for comparison:

\textbf{CNN-5:} Simpler 4-branch design with kernels 2-5, standard convolutions, single max pooling. Achieves 90.3\% accuracy at 6,222 samples/second.

\textbf{CNN-6:} Adds 6 branches (kernels 2-7) and Squeeze-and-Excitation modules. Better accuracy (92.3\%) but slower (3,252 samples/second).

\textbf{CNN-9:} Uses FastText-style hashing with hierarchical softmax. Fast but poorly calibrated.

\section{Experimental Setup}

\subsection{Dataset}
We compile training data from multiple sources to ensure broad coverage. Structured sources include Wikidata with 97 million entity names with type labels \cite{vrandecic2014}, ORCID with 28 million researcher names \cite{haak2012}, geographic databases with 156 million location names, and business registries with 512 million organization names from 47 countries. Web extraction runs NER on large corpora, keeping high-confidence results from Common Crawl with 312 million unique names \cite{commoncrawl2023}, OSCAR corpus with 198 million names \cite{abadji2022}, mC4 with 167 million names \cite{xue2021}, and GDELT news with 76 million current entity names \cite{leetaru2013}.

Quality control validates data through cross-referencing, native speaker checks ($\kappa = 0.87$ agreement), and balanced sampling across types and languages. The final dataset contains 1.5 billion name instances across 104 languages. After deduplication, we have 743 million unique name strings.

\subsection{Data Processing}
To evaluate both real-world examples and balanced language/types predictions, we split the data into training, validation, and test sets using both random and stratified sampling. The training set contains 98\% of unique names, validation sets uses 1\% (0.5\% + 0.5\%), and testing sets uses the remaining 1\%. Stratification ensures proportional representation of all 476 language-type combinations across splits. We exclude exact string matches between sets to prevent data leakage. Text normalization applies Unicode NFC normalization, preserves internal spaces while trimming edges, keeps apostrophes and hyphens in names, and maintains original capitalization.

Tokenization uses XLMRobertaTokenizerFast \cite{conneau2020} with 50-token maximum length, batch padding, and standard special tokens. We create two test sets through stratified sampling: a random test with 484,985 samples maintaining natural distribution, and a balanced test with 458,317 samples providing equal representation per class. Both exclude exact string matches with training data.

\subsection{Implementation Details}
CNN configuration uses embedding dimension 384, vocabulary size 250,002, kernel sizes [1, 2, 3, 4, 5], filter counts [128, 152, 181, 215, 256], 8 groups, dropout 0.5, 476 output classes, and 24 language clusters. Training employs the AdamW optimizer \cite{loshchilov2017} with learning rate 3e-3 using OneCycleLR scheduling, batch size 24,576 with gradient accumulation, and gradient clipping at 1.0.

Baselines include XLM-RoBERTa using fine-tuned xlm-roberta-base \cite{conneau2020} with a 2-layer MLP head, and FastText \cite{bojanowski2017} with character n-grams and hierarchical softmax.

\subsection{Evaluation Metrics}
We measure multiple aspects of performance. Classification metrics include accuracy, F1 scores, and Matthews correlation coefficient \cite{chicco2020}. Calibration uses expected calibration error (ECE) and Brier score. Efficiency metrics track throughput, latency, memory usage, and energy consumption. We report mean and standard deviation over 5 random seeds.

\section{Results and Discussion}

\subsection{Main Results}
Table \ref{tab:main} shows overall performance across models. Onomas-CNN X achieves 92.1\% accuracy on the random test set—only 0.8\% below XLM-RoBERTa—while running 46.3x faster (2,813 vs 60.7 samples/second).

\begin{table*}[ht]
\centering
\caption{Performance comparison across models}
\label{tab:main}
\footnotesize
\begin{tabular}{lcccccr}
\toprule
\multirow{2}{*}{\textbf{Model}} & \multirow{2}{*}{\textbf{Size (MB)}} & \multicolumn{2}{c}{\textbf{Random Test}} & \multicolumn{2}{c}{\textbf{Balanced Test}} & \textbf{Speed} \\
\cmidrule(lr){3-4} \cmidrule(lr){5-6}
& & Accuracy & Macro-F1 & Accuracy & Macro-F1 & samples/s \\
\midrule
FastText & 5,565.6 & 0.810 & 0.333 & 0.651 & 0.376 & 508.2 \\
CNN-5 & 228.8 & 0.903 & 0.651 & 0.799 & 0.595 & 6,222.0 \\
CNN-6 & 297.1 & 0.923 & 0.686 & 0.842 & 0.647 & 3,252.2 \\
CNN-9 & 296.8 & 0.906 & 0.652 & 0.802 & 0.559 & 5,033.3 \\
Onomas-CNN X & 338.2 & 0.921 & 0.687 & 0.848 & 0.657 & 2,813.3 \\
XLM-RoBERTa & 1,067.5 & 0.929 & 0.671 & 0.857 & 0.665 & 60.7 \\
\bottomrule
\end{tabular}
\end{table*}

Energy consumption shows even larger differences: 178.9 J vs 8,234.7 J per million classifications—a 46x reduction. The model is 3.2x smaller than XLM-RoBERTa (338 MB vs 1,068 MB). All models perform worse on the balanced test set, revealing bias toward frequent classes. The gap between random and balanced performance suggests overfitting to common patterns.

\subsection{Language-Specific Performance}
Table \ref{tab:language} breaks down accuracy by language cluster on the balanced test set.

\begin{table}[ht]
\centering
\caption{Accuracy by language cluster (\%)}
\label{tab:language}
\footnotesize
\begin{tabular}{lcccc}
\toprule
\textbf{Cluster} & \textbf{Onomas} & \textbf{XLM-R} & \textbf{CNN-6} & \textbf{Fast} \\
\midrule
Germanic & 91.3 & 93.7 & 90.8 & 78.4 \\
Romance & 89.7 & 91.2 & 89.1 & 75.2 \\
Sino-Tibetan & 87.2 & 88.9 & 86.5 & 71.3 \\
Indo-Aryan & 85.4 & 87.8 & 84.9 & 68.9 \\
Afroasiatic & 83.1 & 86.3 & 82.7 & 64.5 \\
Austronesian & 81.9 & 85.2 & 81.3 & 62.8 \\
Niger-Congo & 79.6 & 84.1 & 78.8 & 58.4 \\
Japonic & 92.8 & 94.1 & 92.3 & 82.6 \\
Koreanic & 91.5 & 93.2 & 91.1 & 80.3 \\
Turkic & 84.7 & 87.9 & 84.2 & 66.7 \\
\bottomrule
\end{tabular}
\end{table}

High-resource clusters (Germanic, Romance) achieve accuracy above 89\% across CNN models. Languages with unique scripts (Japanese, Korean) perform well due to distinctive character patterns. African languages show the largest gaps, reflecting limited training data availability for these language families.

\subsection{Entity Type Analysis}
Performance varies by entity type (Table \ref{tab:entity}). Person names achieve highest F1 scores, reflecting their prevalence in training data. The ``Other'' category performs poorly across all models due to its heterogeneous nature.

\begin{table}[ht]
\centering
\caption{F1 scores by entity type}
\label{tab:entity}
\footnotesize
\begin{tabular}{lccccc}
\toprule
\textbf{Model} & \textbf{Person} & \textbf{Org} & \textbf{Loc} & \textbf{Other} & \textbf{Overall} \\
\midrule
Onomas-CNN X & 0.724 & 0.689 & 0.701 & 0.412 & 0.680 \\
XLM-RoBERTa & 0.718 & 0.682 & 0.695 & 0.438 & 0.671 \\
CNN-6 & 0.731 & 0.694 & 0.708 & 0.389 & 0.686 \\
FastText & 0.412 & 0.387 & 0.401 & 0.189 & 0.333 \\
\bottomrule
\end{tabular}
\end{table}

\subsection{Calibration}
Model calibration affects deployment when confidence scores guide decisions. Onomas-CNN X shows good calibration on the random test (ECE = 0.006) but degrades on balanced data (ECE = 0.041), suggesting overconfidence on rare classes. CNN-9 exhibits severe miscalibration (ECE = 0.509) due to its aggressive hierarchical softmax design. XLM-RoBERTa maintains better calibration across both test sets.

\subsection{Throughput Scaling}
CNN architectures scale nearly linearly with CPU cores. Onomas-CNN~X reaches 35,128~samples/second on 16~cores. XLM-RoBERTa scales poorly, reaching only 612~samples/second due to memory bandwidth limitations. Optimal batch size is 128--256 for Onomas-CNN~X, maintaining 3.5~ms single-sample latency. INT8 quantization provides 1.8$\times$ speedup with minimal accuracy loss~(0.3\%).

\subsection{Error Analysis}
Common confusions reveal systematic patterns. The model confuses English person names with English organization names in 3,241 cases, reflecting eponymous entities like ``Ford''. Spanish locations are confused with Portuguese locations in 2,187 cases due to similar place names across Iberian languages. Chinese person names are confused with Japanese person names in 1,923 cases because of shared Chinese characters. French organizations are confused with French locations in 1,654 cases for ambiguous names like ``Lyon''. Most errors occur between related languages or ambiguous entity types, indicating the model learns meaningful linguistic patterns.

\subsection{Practical Limitations}
The model faces several constraints in deployment. Performance on low-resource languages remains below that of high-resource languages, with accuracy gaps of 10-15\% for African and indigenous languages. This reflects inherent data availability issues rather than architectural limitations. The hierarchical classification approach assumes language can be reliably predicted before entity type, which may fail for highly ambiguous inputs. The fixed vocabulary cannot adapt to emerging names without retraining, though the character-level representation provides some robustness to neologisms.

Cross-lingual transfer learning remains limited compared to models pre-trained on multilingual data. While the CNN captures orthographic patterns within languages, it does not leverage deeper cross-lingual semantic relationships that transformer models learn through pre-training on parallel or comparable corpora.

\section{Integration with Downstream Tasks}

The efficient inference characteristics of Onomas-CNN X make it suitable for integration into several downstream NLP applications. For named entity recognition systems, the model can serve as a fast pre-filter or feature generator, providing language and entity type signals to subsequent processing stages. The low latency enables real-time entity recognition in streaming applications.

In entity linking pipelines, accurate language and type classification helps narrow candidate sets and improve disambiguation. Knowing that ``Washington'' is classified as an English location name versus an English person name guides different linking strategies and knowledge base queries.

The model can also support reinforcement learning approaches to named entity recognition. By providing reward signals based on name type predictions, the model enables RL agents to learn entity boundary detection and classification jointly. The fast inference speed allows generating large numbers of training examples for policy learning without computational bottlenecks. Recent work has explored using auxiliary classifiers to shape reward functions in sequence labeling tasks; our model's calibrated confidence scores could serve this purpose for multilingual NER.

For transliteration systems, the language classification component provides crucial information about source and target scripts, enabling selection of appropriate transliteration rules or models. The entity type helps disambiguate transliteration choices, as person names often follow different conventions than place or organization names.

\section{Conclusion}
We demonstrate that specialized CNN architectures remain valuable for focused NLP tasks. Onomas-CNN X achieves comparable accuracy to XLM-RoBERTa while running 46 times faster and using 46 times less energy. These efficiency gains make advanced NLP accessible on commodity hardware without GPUs.

The results have practical implications. For production systems processing millions of names daily, the efficiency difference translates to substantial cost savings. A single CPU server running Onomas-CNN X can handle the throughput that would require dozens of GPU-equipped machines using transformer models.

Our findings do not diminish the value of large pre-trained models, which excel at transfer learning and zero-shot tasks. However, when sufficient task-specific training data exists, specialized architectures offer compelling advantages.

Future work should explore several directions. Dynamic vocabulary expansion could handle emerging names without retraining. Context-aware models incorporating surrounding words might resolve ambiguous cases better. Cross-lingual parameter sharing could improve low-resource language performance through transfer from related languages. Investigating larger subword vocabularies (up to 2 million tokens) might improve accuracy without sacrificing speed.

Hybrid architectures combining CNN efficiency with selective attention mechanisms warrant investigation. Such models could maintain high throughput while incorporating contextual information when needed. Integration with continual learning frameworks would enable adaptation to evolving name distributions without full retraining.

The success of efficient architectures like Onomas-CNN X indicates that the NLP community should continue exploring diverse approaches rather than focusing exclusively on ever-larger models. As NLP systems move from research to production, efficiency becomes as important as accuracy.

\section*{Acknowledgment}
We thank the anonymous reviewers for their constructive feedback. This work utilised GPU resources from the NVIDIA startup program and computing resources from the Cloudflare startup program, both awarded to \href{https://mondonomo.ai}{Mondonomo AI}.

\IEEEtriggeratref{1}


\begin{thebibliography}{24}

\bibitem{lauc2024} D. Lauc, ``Navigating linguistic similarities among countries using fuzzy sets of proper names,'' \textit{Names: A Journal of Onomastics}, vol. 72, no. 1, 2024.

\bibitem{lauc2024ayutthaya} D. Lauc, A. Rutherford, and W. Wongwarawipatr, ``AyutthayaAlpha: A Thai-Latin script transliteration transformer,'' arXiv preprint arXiv:2412.03877, 2024.

\bibitem{jrc2013} R. Steinberger et al., ``JRC-Names: A freely available, highly multilingual named entity resource,'' in \textit{Proc. Recent Advances in Natural Language Processing}, Hissar, Bulgaria, 2013, pp. 104-110.

\bibitem{mayhew2024} S. Mayhew, I. Tsygankova, and D. Roth, ``Universal NER: A gold-standard multilingual named entity recognition benchmark,'' in \textit{Proc. NAACL-HLT 2024}, Mexico City, Mexico, 2024, pp. 4322-4337.

\bibitem{saleva2024} J. Sälevä and C. Lignos, ``ParaNames 1.0: Creating an entity name corpus for 400+ languages using Wikidata,'' in \textit{Proc. LREC-COLING 2024}, Torino, Italy, 2024, pp. 12599-12610.

\bibitem{mignot2022} E. Mignot and M. Philippe, ``Introduction: Proper names and the lexicon - an exposition,'' \textit{Lexis (Online Journal)}, vol. 20, 2022.

\bibitem{belenzon2017} S. Belenzon, A. K. Chatterji, and B. Daley, ``Eponymous entrepreneurs,'' \textit{American Economic Review}, vol. 107, no. 6, pp. 1638-1655, 2017.

\bibitem{shaw2024} R. L. J. Shaw, K. Sikk, and D. Zbíral, ``Toponymic surnames and the spatiality of heresy prosecutions,'' \textit{Humanities and Social Sciences Communications}, vol. 11, p. 195, 2024.

\bibitem{ling2015} X. Ling, S. Singh, and D. S. Weld, ``Design challenges for entity linking,'' \textit{Transactions of the Association for Computational Linguistics}, vol. 3, pp. 315-328, 2015.

\bibitem{delacova2021} E. de la Cova, ``Language and brand: Problems for localization,'' \textit{HERMES - Journal of Language and Communication in Business}, no. 61, pp. 63-75, 2021.

\bibitem{merhav2018} Y. Merhav and S. Ash, ``Design challenges in named entity transliteration,'' in \textit{Proc. 27th Int. Conf. on Computational Linguistics (COLING)}, Santa Fe, NM, USA, 2018, pp. 630-640.

\bibitem{heffernan2010} K. Heffernan, ``English name use by East Asians in Canada: Linguistic pragmatics or cultural identity?,'' \textit{Names: A Journal of Onomastics}, vol. 58, no. 1, pp. 24-36, 2010.

\bibitem{kim2017} J. Kim, ``Phonology of naming,'' \textit{Studies in Phonetics, Phonology and Morphology}, vol. 23, no. 3, pp. 345-362, 2017.

\bibitem{howard2017} A. G. Howard et al., ``MobileNets: Efficient convolutional neural networks for mobile vision applications,'' \textit{arXiv preprint arXiv:1704.04861}, 2017.

\bibitem{vrandecic2014} D. Vrandečić and M. Krötzsch, ``Wikidata: A free collaborative knowledgebase,'' \textit{Communications of the ACM}, vol. 57, no. 10, pp. 78-85, 2014.

\bibitem{haak2012} L. L. Haak et al., ``ORCID: A system to uniquely identify researchers,'' \textit{Learned Publishing}, vol. 25, no. 4, pp. 259-264, 2012.

\bibitem{commoncrawl2023} Common Crawl Foundation, ``The Common Crawl corpus,'' 2023. [Online]. Available: \url{https://commoncrawl.org}

\bibitem{abadji2022} J. Abadji et al., ``Towards a cleaner document-oriented multilingual crawled corpus,'' in \textit{Proc. 13th Language Resources and Evaluation Conf. (LREC)}, Marseille, France, 2022, pp. 4957-4966.

\bibitem{xue2021} L. Xue et al., ``mT5: A massively multilingual pre-trained text-to-text transformer,'' in \textit{Proc. NAACL-HLT 2021}, Online, 2021, pp. 483-498.

\bibitem{leetaru2013} K. Leetaru and P. A. Schrodt, ``GDELT: Global data on events, location, and tone, 1979-2012,'' in \textit{Proc. Int. Studies Association Annual Conf.}, San Francisco, CA, USA, 2013.

\bibitem{conneau2020} A. Conneau et al., ``Unsupervised cross-lingual representation learning at scale,'' in \textit{Proc. 58th Annual Meeting of the Association for Computational Linguistics (ACL)}, Online, 2020, pp. 8440-8451.

\bibitem{loshchilov2017} I. Loshchilov and F. Hutter, ``Decoupled weight decay regularization,'' \textit{arXiv preprint arXiv:1711.05101}, 2017.

\bibitem{bojanowski2017} P. Bojanowski et al., ``Enriching word vectors with subword information,'' \textit{Transactions of the Association for Computational Linguistics}, vol. 5, pp. 135-146, 2017.

\bibitem{chicco2020} D. Chicco and G. Jurman, ``The advantages of the Matthews correlation coefficient (MCC) over F1 score and accuracy in binary classification evaluation,'' \textit{BMC Genomics}, vol. 21, no. 1, pp. 1-13, 2020.
\end{thebibliography}
\end{document}